\documentclass[10pt,twocolumn,letterpaper]{article}

\usepackage{cvpr}              

\usepackage{graphicx}
\usepackage{amsmath}
\usepackage{amssymb}
\usepackage{booktabs}
\usepackage{xcolor}

\usepackage{tcolorbox}
\usepackage{fancyvrb}
\usepackage[accsupp]{axessibility}

\DefineVerbatimEnvironment{modeloutput}{Verbatim}{
	frame=single,
	framesep=2mm,
	fontsize=\small
}

\newenvironment{llmprompt}[1]{%
    \section*{Prompt: #1}%
    \begin{quote}%
}{%
    \end{quote}%
}

\usepackage[pagebackref,breaklinks,colorlinks]{hyperref}

\usepackage[capitalize]{cleveref}
\crefname{section}{Sec.}{Secs.}
\Crefname{section}{Section}{Sections}
\Crefname{table}{Table}{Tables}
\crefname{table}{Tab.}{Tabs.}

\def\cvprPaperID{42553} 
\def\confName{CVPR}
\def\confYear{2026}

\newcommand{\note}[1]{}

\begin{document}

\title{Bootstrapping Sign Language Annotations \\ with Sign Language Models}

\author{Colin Lea\\
Apple\\
{\tt\small colin\_lea@apple.com}
\and
Vasileios Baltatzis\\
Apple\\
{\tt\small vbaltatzis@apple.com}
\and
Connor Gillis\\
Apple\\
{\tt\small connorgillis@apple.com}
\and
Raja Kushalnagar$^{*}$\\
Gallaudet University\\
{\tt\small raja.kushalnagar@gallaudet.edu}
\and
Lorna Quandt$^{*}$\\
Gallaudet University\\
{\tt\small lorna.quandt@gallaudet.edu}
\and
Leah Findlater\\
Apple\\
{\tt\small lfindlater@apple.com}
}
\maketitle
\renewcommand{\thefootnote}{*}
\footnotetext{Work done entirely at Apple.}
\renewcommand{\thefootnote}{\arabic{footnote}}

\begin{abstract}
AI-driven sign language interpretation is limited by a lack of high-quality annotated data. 
New datasets including ASL STEM Wiki and FLEURS-ASL contain professional interpreters and 100s of hours of data but remain only partially annotated and thus underutilized, in part due to the prohibitive costs of annotating at this scale. 
In this work, we develop a pseudo-annotation pipeline that takes signed video and English as input and outputs a ranked set of likely annotations, including time intervals, for glosses, fingerspelled words, and sign classifiers. 
Our pipeline uses sparse predictions from our fingerspelling recognizer and isolated sign recognizer (ISR), along with a K-Shot LLM approach, to estimate these annotations. 
In service of this pipeline, we establish simple yet effective baseline fingerspelling and ISR models, achieving state-of-the-art on FSBoard (6.7\% CER) and on ASL Citizen datasets (74\% top-1 accuracy). 
To validate and provide a gold-standard benchmark, a professional interpreter annotated nearly 500 videos from ASL STEM Wiki with sequence-level gloss labels containing glosses, classifiers, and fingerspelling signs. 
These human annotations and over 300 hours of pseudo-annotations are being released in supplemental material.
\end{abstract}

\section{Introduction}
\label{sec:intro}

Accurate systems for automated sign language recognition have the potential to reduce barriers and improve access to technology for the millions of people who use a sign language (SL) as a primary language, along with non-signers trying to communicate with signers.
Hearing people routinely benefit from voice-based interaction on smart devices and for services like automated captions; unfortunately, these systems are inadequate for many people who are deaf or hard of hearing (DHH)~\cite{glasser2021understanding}.
If high-quality sign recognition technology existed, many deaf people would prefer to communicate in sign language, allowing them to communicate in their first or preferred language ~\cite{bragg_sign_2019,braggassets23}. 

The best performing sign language recognition models use glosses as an intermediate representation of sign~\cite{Peike_NeurIPS2025}. 
Glosses are the closest equivalent of written text (\eg English word) for a given sign (\eg the sign for a feline pet may be glossed as CAT).\footnote{We adopt textual glosses, but acknowledge other forms exist, such as HamNoSys~\cite{hanke2004hamnosys} and Sign Writing~\cite{sutton2010signwriting}} 
Glosses are inherently lossy, as a given sign is often more complex than the gloss may suggest, but they offer a way to standardize approximately what each sign means and represent the grammatical structure of a sentence. 
In addition to individual glosses, most sign languages can also be notated using fingerspelling, classifiers, name signs, lexicalized fingerspelling, and pointing cues to refer to the spatial location of entities being discussed~\cite{valli2000linguistics}.
Some recent systems use gloss-free approaches that go directly from video to written text (\eg~\cite{SINCAN2025104498,Zhou_2023_ICCV,chen-etal-2024-factorized,JangLostInTranslation_CVPR2025}).
However, even when using 3,000 hours of YouTube-quality signing video, 
current gloss-free approaches are insufficient for training generalizable and robust systems~\cite{tanzer2025youtubesl}.

Glosses are important in part due to large differences in grammar and linguistic structures between signed and spoken languages, but they are costly to annotate. 
Annotating sign language data requires sign language and linguistics expertise and is very time-consuming, with estimates ranging from 20 seconds to annotate one second of video for our glossing convention, to 30 seconds for multidimensional annotations using ELAN~\cite{dreuw2008towards}. 
The motivating hypothesis for our work is that if we can accurately recognize some of the structures from a signed video, namely fingerspelled words and a predefined set of known signs from an isolated sign recognition model, then we can fill in candidate out-of-domain glosses, and use them for downstream recognition tasks or as initialization for human annotations.
Our contribution is a system, not a novel architecture: we focus on scalably generating pseudo-annotations to address the critical bottleneck of large, high-quality annotated datasets with diverse signing patterns.

Our pseudo-annotation pipeline has three steps: (1) given an English sentence, generate $k$ plausible glossings using LLM prompting, (2) for each candidate gloss sequence, detect which words are fingerspelled and align in time with the video, and (3) use an isolated sign recognition model to detect if each gloss was likely signed.
To develop this system, we built state-of-the-art models for ASL fingerspelling (letters, numbers, symbols) and Isolated Sign Recognition (ISR) using a vocabulary of 2,731 glosses by leveraging the FSBoard~\cite{FSBOARD} and ASL Citizen~\cite{desai2023asl} datasets, respectively. 
The output of the pipeline is time-aligned annotations, including segments corresponding to fingerspelled words, known glosses within the sign vocabulary, and new/unknown glosses that the model has not seen before. 
Each candidate sequence is ranked using an aggregate score from these models. 

We validate this approach using new annotations we created and will publicly release for nearly 500 videos from ASL STEM Wiki~\cite{ASL_STEM_Wiki}, where an expert interpreter manually glossed each sentence using the video as the primary reference and English text as the secondary reference. 
Furthermore, we use this pipeline to enhance the ASL STEM Wiki~\cite{ASL_STEM_Wiki} and FLEURS-ASL~\cite{tanzer-2025-fleurs} datasets, each of which provides high-quality video interpretations of content from Wikipedia, containing ASL video and English text for each phrase, but no glosses.
These pseudo-annotations are also being released publicly. 

Our contributions are as follows
\begin{itemize}
    \item A new pseudo-annotation pipeline that takes text and ASL video and generates sign language annotations.
    \item Model architectures that achieve state of the art performance on isolated sign recognition on ASL Citizen and fingerspelling recognition on FSBoard.
    \item Development of nearly 500 manual English-to-gloss annotations and validation through back translation.
    \item Public release of these manual annotations and pseudo-annotations for over 300 hours of ASL STEM Wiki and 7.5 hours of FLEURS-ASL.
\end{itemize}






\section{Related Work}
\label{sec:related}

Systems for sign language recognition have benefited from recent advancements in machine learning.
Some of the more popular datasets driving research on sign language interpretation include
PHOENIX-2014T (German Sign Language)~\cite{camgoz2018phoenix}, 
BOBSL (British Sign Language)~\cite{albadrashiny2022bobsl} , 
CSL-daily (Chinese Sign language)~\cite{zhou2021improving}, 
and How2Sign (American Sign Language)~\cite{duarte2021how2sign}.
PHOENIX-2014T, BOBSL, and CSL-daily all have some gloss annotations and generally models perform significantly better when using them. 
To contrast, we are not aware of any continuous ASL recognition dataset at this scale with annotated glosses. 
New datasets such as ASL STEM Wiki~\cite{ASL_STEM_Wiki}, OpenASL~\cite{shi2022open}, YouTube-ASL~\cite{tanzer2025youtubesl}, and FLEURS ASL~\cite{tanzer-2025-fleurs} offer hundreds of hours of ASL signing, but only contain English transcripts or captions. 

There are two recent sign language papers focused generating pseudo glosses. Sign2GPT~\cite{wong_sign2gpt_2024} leverages linguistics information, like part-of-speech tagging, to translate between English and ASL glosses. 
Unlike us, they assume a single candidate translation between languages, which is based on English order, and which does not account for the many possible differences in grammar and style. 
The pseudo annotation approach from Guo \etal \cite{Peike_NeurIPS2025} uses a weakly supervised approach, first generating a candidate glossing, and then swapping the sign order using their model. 
Unlike our approach, they do not incorporate fingerspelling and they only fix the set of candidate glosses 

Other work has focused on reducing reliance on glosses by going directly from video to written text, often using self-supervised or weakly-supervised learning(\eg~\cite{SINCAN2025104498,Zhou_2023_ICCV,chen-etal-2024-factorized,JangLostInTranslation_CVPR2025,zuo2023cico}).
Our work complements these efforts by providing a method to create dense annotations required for fully-supervised training and evaluation, specifically for American Sign Language (ASL).
Our annotations can serve as weak supervision in gloss-based training or as intermediate supervision for gloss-free approaches.



Our pipeline builds on recent work in fingerspelling recognition, isolated sign recognition, and LLM-based translation. 
For typical discourse in ASL, around 15\% of words are fingerspelled \cite{padden2003alphabet,keane2016fingerspelling}. However, many existing continuous sign language recognition systems do not explicitly model letters or numbers. 
One exception is Varol \etal~\cite{varol2021gaslr} who unified models that can decode both glosses and fingerspelled characters.
Historically, fingerspelling datasets were small, but with the release of FSBoard~\cite{FSBOARD}, there are now over 250 hours of high-quality ASL fingerspelling data with per-phrase labels with letters, numbers, and symbols. 
In this work, we develop a state-of-the-art model on the FSBoard benchmark and demonstrate its efficacy on videos with both fingerspelling and other signs. 

\begin{figure*}[t]
    \centering
    \includegraphics[width=0.95\textwidth]{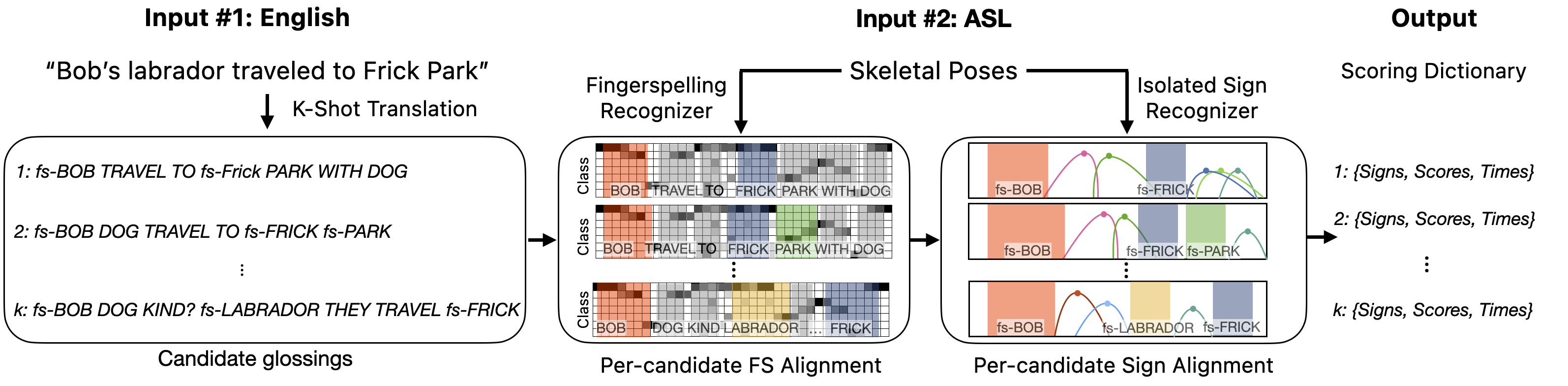}
    \caption{Our pseudo-annotation pipeline takes in English text and ASL video and outputs likely gloss annotations. There are three steps: K-Shot Candidate Translation, Fingerspelling Alignment, and Sign Alignment.}
    \label{fig:pseudo_pipeline}
\end{figure*}

Isolated Sign Recognition (ISR) is the classification of pre-segmented individual signs, generally from a pre-determined dictionary of signs labels. 
Some recent ISR models build on video understanding architectures such as I3D~\cite{carreira2017i3d}, SlowFast~\cite{feichtenhofer2019slowfast}, and MViT~\cite{fan2021mvit}. 
There has been a push to larger vocabularies including WLASL~\cite{li2020wlasl}, which is a YouTube-based ISR dataset with 2000 sign classes, and ASL Citizen~\cite{desai2023asl}, which contains video from consenting signers with 2731 sign classes. 
Our contribution to this area is the development of a new state-of-the-art model on ASL Citizen~\cite{desai2023asl} and application within our broader pipeline.

Lastly, LLMs have been used for prompt-based text-to-gloss and gloss-to-text translation (\eg ~\cite{Mercanoglu_Sincan_2025,zhang2025towards,Peike_NeurIPS2025} and for LLM-tuned systems (\eg SignLLM~\cite{jiang2024signllm}).
In this work, we use pre-trained LLMs such as Claude Sonnet 4.5 to generate batches of candidate gloss translations with glossing-specific prompting strategies





\section{Pseudo-annotation Pipeline}
\label{sec:pipeline}

Our goal is to take an English sentence and a corresponding video clip and generate a set of likely sign language annotations. 
We developed a pipeline, depicted in Figure~\ref{fig:pseudo_pipeline}, which starts off with many candidate translations, and narrows down the selection based on subsequent models. 
First, an LLM prompt with an input English sentence generates $k$($=10$) candidate translations, where each translation may have distinct grammar or style.
Next, a fingerspelling recognizer with a video and candidate translation returns a list of terms likely fingerspelled, along with time spans. 
Finally, a sign alignment and scoring model using the isolated sign recognizer computes the likelihood of each translation being correct, and identifies at what frame each sign occurs.
The result is a per-translation score with time-stamps and scores for each element in a translation. 

\label{sec:gloss_conventions}
Models and annotations throughout this work use the following glossing conventions:
    
    \noindent\textbf{Glosses} are written in uppercase (\eg a cat is CAT). 
    
    \noindent\textbf{Fingerspelling} is notated \texttt{fs-WORD} and can include letters, numbers, or special characters (\eg \texttt{fs-12.34}). Notation \texttt{fs-} is applied per word (\eg \texttt{fs-NEW fs-YORK}).
    
    \noindent\textbf{Name signs} are notated with \texttt{ns-NAME} to indicate that there is a custom sign for that entity (\eg City names).
    
    \noindent \textbf{Lexicalized fingerspelling} and fingerspelled abbreviations are defined using \texttt{\#WORD} notation to indicate that hand shapes or motions differ from the orthographic definition (\eg \texttt{\#BANK} blends \texttt{B-N-K} and omits \texttt{A}; \texttt{CORPORATION} is frequently fingerspelled \texttt{CORP}, which we represent as \texttt{\#CORPORATION}).
    
    \noindent \textbf{Classifiers} use convention \texttt{CL:X(TEXT)} where \texttt{X} is a classifier hand shape and \texttt{TEXT} describes the motion. (\eg the sequence \texttt{CL:4(list) CL:1(point finger) ITEM} indicates the user creating a 4-item list (\ie with their non-dominant hand), points to a finger, and then signs the gloss for that item.)

\subsection{LLM Translation Candidates}

Given the low quantity of paired ASL glosses and English, we follow recent work on low-resource machine translation (\eg \cite{iyer-etal-2024-exploring}) and rely on LLM prompting to generate plausible translations between English and ASL glosses. 
Recall the grammatical structure of many signed and written text pairs (\eg ASL and English) differ linguistically and from signer-to-signer. 
For instance, a native signer of ASL who learned from birth will have a markedly different signing style than a hearing person who acquired ASL as an adult after using English their whole life~\cite{hilger2015second,schonstrom2022l2m1}. 
As such, we prompt the LLM to output $k$ candidate translations that represent different ways in which someone might have interpreted a given piece of text to sign, taking into account stylistic- or skills- based differences. 

Our prompt, shown in full in the supplemental material, encourages the model to use rules derived from the gloss conventions described above. 
It includes rules for numbers, fingerspelling, grammar, use of classifiers, and other common convention in ASL like references to time.
Additional rules were explored, inclusion of the ISR vocabulary, but ultimately the LLM appeared to ignore these suggestions, and thus they are not included the in final prompt. 
The output is a json string containing $k$ glossed sentences corresponding to the English input.





\begin{figure}[t!]
    \centering
    \includegraphics[width=0.5\textwidth]{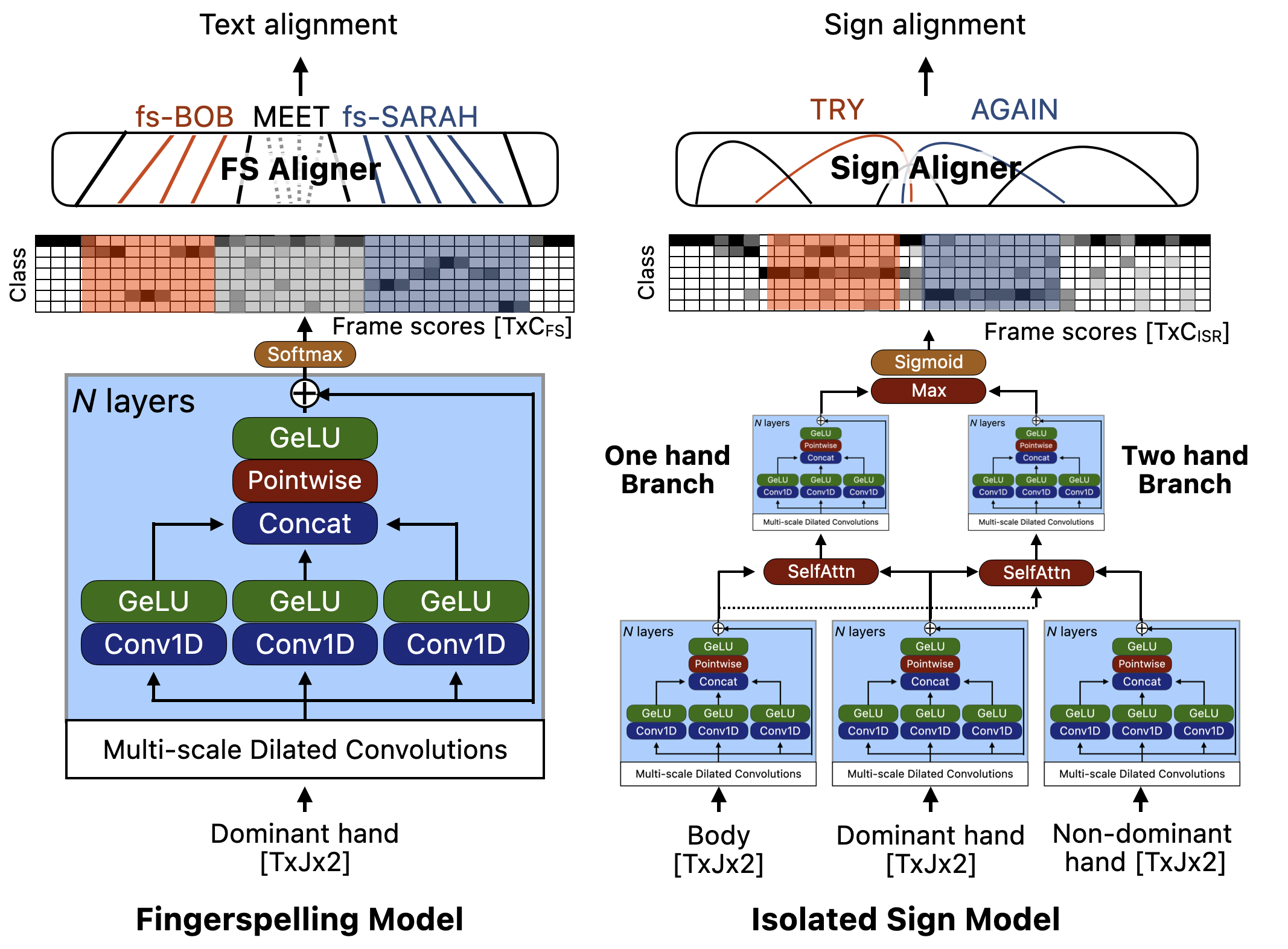}
    \caption{(left) Architecture and design of the (left) Fingerspelling and (right) Isolated Sign Recognition models. }
    \label{fig:models}
\end{figure}

\subsection{Fingerspelling Recognizer}
\label{sec:fingerspelling_model}

In fluent signing, fingerspelling tends to be very quick. 
When letters are shown in isolation, most are inherently static while in ASL $j$ and $z$ are dynamic. 
In practice, there is significant coarticulation and some letters may be blended or omitted entirely (\eg `acid` may be signed with a blended c/i shape in the middle).
Thus, it is important to be able to detect both static representations and coarticulations. 

Our fingerspelling model consists of two parts: a module that outputs per-frame probabilities of each character and a module that detects or aligns words that appear in a sequence.
A key design principle for this model is the ability to process arbitrarily long input sequences efficiently with high resolution temporal outputs for time stamping purposes.
The model, depicted in Figure~\ref{fig:models}(left), is a Multi-scale Temporal Convolutional Network (TCN) with dilated convolutions and was chosen over Transformer- or Conformer-based approaches based on performance in early experiments and their improved ability to provide accurate timestamps.
See the supplemental material for a comparison between TCN and Conformer architectures demonstrating the TCN's superior cross-dataset generalization.
The TCN comprises of multiple blocks (\eg $N=6$) with residual connections, where each block contains dilated convolutions with different strides and GeLU activations, which are then concatenated. 
Dilation rates increase progressively across layers to widen the receptive field so at any given time the model sees several fingerspelled letters.
The final TCN layer outputs frame-wise features, which are then projected via a linear layer to produce logits for the 26 English alphabet characters, numbers 0-9, special symbols, and a blank token. 



The model input is a sequence of 21 2D skeletal joints extracted using MediaPipe's pose estimator~\cite{lugaresi2019mediapipe} from the dominant hand. 
Hand coordinates are normalized relative to the 2D mean of the hand and using the standard deviation per joint and flattened into a 42-dimensional feature vector per frame. 
Missing frames are imputed using joint positions from the nearest valid frame. 
Input and output sample rates are synchronized with the video frame rate, typically 30 FPS, though ranging from 24 to 60 FPS across videos. 
The dominant hand is selected based on the percentage of valid hand frames per video. 

The model is trained using the Connectionist Temporal Classification (CTC) loss~\cite{graves2006connectionist} where each video is weakly annotated with the intended fingerspelled phrase. 
The tokens are formatted with notation \texttt{\{| C A T | A N D | D O G |\}} where each character is separated with a space and each word is separated with a pipe (\texttt{|}) symbol.
In practice, on the FSBoard dataset, there are a non-trivial number of videos where the signed phrase differs from the intended phrase. For example, someone starts fingerspelling a word, stops mid-way, and then re-spells the word. 
Some videos also have poor hand tracking. 
Both video types have poor losses and samples are automatically removed as part of a two-pass process in training the model.

The fingerspelling model has about 2M parameters with a receptive field of 35 frames and took approximately 2 hours to train to 100 epochs on one Nvidia A100 GPU.
Training uses AdamW with learning rate 1e-4, weight decay 1e-5, and cosine annealing. We use a 70/15/15 train/val/test split stratified by signer with early stopping patience of 10 epochs.

\noindent \textbf{Recognition \& Alignment:} The fingerspelling model is used in two ways: (1) for context-free \textit{fingerspelling recognition}, where the input video consists solely of someone fingerspelling (\eg the typical test case with FSBoard), and (2) for \textit{fingerspelling alignment} as part of our pipeline where the input is both ASL video, which may contain fingerspelling and other signing elements, along with text corresponding to the English transcript or a candidate glossing. 

For fingerspelling recognition, our baseline approach uses a simple greedy decoder to output a text sequence, taking the highest probability class at each frame, merging each repeated letter, and replacing pipes with spaces (\eg \texttt{\{| AAA BB | CCCC |\} $\rightarrow$ \{AB C\}}).
While it may be advantageous to jointly train this model with an LLM, this baseline already surpasses state of the art. In addition, we show that using the raw output with a stand alone LLM using an error-correcting prompt is sufficient for fixing the majority of formatting errors.
See the supplemental material for the prompt. 

For fingerspelling alignment, we compute forced alignment~\cite{gorman2011prosodylab} between the text (\ie a sentence or glossing) and the per-frame fingerspelling model probabilities, using the raw character-level text as input. 
The output of this function is a probability and timestamp for each character. 
These scores are aggregated so that each word from the text is parameterized by it's average score and time interval.
While perhaps counterintuitive, in Section~\ref{sec:experiments} we show that the model is very effective at distinguishing between words, even when using raw English sentences as input. 




\subsection{Isolated Sign Recognizer}
\label{ssec:isolated_sign_model}

The isolated sign recognizer takes in a sequence of skeletal poses and, for each frame, predicts (a) if there is any sign being performed and (b) for each of the 2731 signs in our vocabulary if that sign is being performed. Its architecture, illustrated in Figure~\ref{fig:models}(right), uses the same convolutional modules as the fingerspelling model but with two non-standard elements. 
First, we identify whether the person is likely to be signing in a manner that is left-handed or right-handed, and we flip the model so that in either case the dominant-hand is presented as the right. 
Results with both default and ideal handedness are shown in Section~\ref{sec:experiments}. 
Second, we learned that our earlier models sometimes learned spurious correlations between the hands during signs that rely only on one hand. 
This is likely an artifact of dataset size. 
Thus, our architecture has two internal branches: one that takes in body, dominant, and non-dominant hands, and another that only takes the body and dominant hand. 
At the end of the network we take the max of the logits per-class between the two-hand and one-hand branches.

The model input is the sequence of upper body 2D skeletal joint coordinates from MediaPipe's ``holistic'' pose estimator (torso, legs, arms, and partial hands and face) and the dominant and non-dominant hand. 
Holistic pose is normalized using the center of the shoulders and shoulder width and the hands are normalized the same as with fingerspelling. We remove joints below the hips.

The ISR model is trained using binary cross entropy per-frame and per-class with additional classes for `any' sign and a null class.
The start and end times of each clip are augmented by up to 25\% on each side, and for efficiency purposes, batches are concatenated in time. 
The final model has 6.6M parameters, a receptive field of 23 frames, and took $\sim$2 hours to train to 200 epochs on one A100 GPU.
Training uses AdamW with learning rate 1e-4, weight decay 1e-5, and cosine annealing, with early stopping patience of 10 epochs. Data splits follow the same 70/15/15 stratification by signer as the fingerspelling model.

\noindent \textbf{Recognition \& Alignment:}
The ISR model is used in two ways. 
For sign classification tasks (\ie the standard setting for ASL Citizen), the model is run on a given video and the output is the highest probability non-null class when max pooled across time. 
Within our pipeline, the ISR model is used to detect if and when a gloss is signed in a video using the gloss candidates and fingerspelling detections as input. 
One challenge is that many potential glosses are out of vocabulary, and thus our model is unable to recognize them. 
For these we use the \texttt{ANY} class from the ISR model to estimate the likelihood of a sign being performed. 
Similar to fingerspelling, we use forced alignment on each interval between detected fingerspelling words. 
Using the example \texttt{fs-BOB TRAVEL TO fs-FRICK PARK WITH DOG} from Figure~\ref{fig:pseudo_pipeline}, alignment is computed for \texttt{TRAVEL TO} between \texttt{fs-BOB} and \texttt{fs-FRICK} and alignment of glosses \texttt{PARK WITH DOG} between \texttt{fs-FRICK} and the end of the video. 


\subsection{Candidate Rankings}
For each $k$ candidate translation, we compute a score $ \sum_{g \in G} S(g, a)$ where $g$ is a sign, $a$ is the alignment, $G$ is the set of signs for a translation, and scoring function $S(g,a)$ corresponds to the fingerspelling or gloss score as computed using their respective alignment functions.
In other words, candidates are ranked by the average likelihood of each sign in a sequence, reflecting how many signs in the candidate are detected by the fingerspelling and sign recognition models.
Candidates are then ranked based on this score. 

The final output of the system is the set of candidate translations, aggregate scores for each translation, likelihood of each gloss being fingerspelled or a sign, and a time interval or frame for each gloss. 
It is assumed that in-vocabulary signs are likely to be signed as the gloss implies, though out-of-vocabulary signs may correspond to the given sign or a different but similar sign. 
\section{Data \& Annotations}
\label{sec:data}

In this section, we describe the two datasets we use for pseudo-annotations, details on our manual annotation process, and datasets we use for fingerspelling and ISR. 

\subsection{Pseudo Annotation Datasets}

ASL STEM Wiki~\cite{ASL_STEM_Wiki} is a specialized dataset focused on technical vocabulary from Science, Technology, Engineering, and Mathematics, which was created to address the lack of technical signs in most general-purpose ASL datasets. 
It features 37 professional ASL interpreters, who interpret Wikipedia pages corresponding to STEM concepts, such as ``algorithm,'' ``mitochondria,'' or ``calculus.''
Each video is annotated with the English sentence being interpreted, along with metadata about the high-level Wikipedia topic. 
Additionally, there is a 500-video subset with fingerspelling annotations.
We use this dataset because it contains consenting signers, is large in size, and has high quality video.

FLEURS-ASL~\cite{tanzer-2025-fleurs} is a 1,749-video American Sign Language component of the otherwise spoken language FLEURS dataset~\cite{conneau2023fleurs}, with 7.5 hours of content. 
FLEURS contains parallel speech and text from over 100 languages and is also drawn from Wikipedia articles. 
The ASL portion consists of professional studio recordings of ASL signers translating a standardized set of English sentences that are common across all languages in the benchmark.
Similar to ASL STEM Wiki, each video is annotated with the English sentence being interpreted. 
An additional set of annotations FLEURS-ASL-FS~\cite{tanzer2024fingerspellingsignlanguagetranslation} contains annotations indicating which words in a sentence are fingerspelled. 
We use this dataset because it contains consenting signers, high quality video, and is relevant for low-resource translation. 

Note that we do not use the popular
How2Sign dataset~\cite{duarte2021how2sign} based on findings of Tanzer \etal~\cite{tanzer2024reconsideringsentencelevelsignlanguage}, whose work with human graders showed that a large portion of videos are too ambiguous for many fluent signers to properly translate.
We exclusively use datasets with consenting signers; licensing and ethical considerations precluded web-scraped sources like YouTube-ASL.

\subsection{ASL STEM Wiki Gloss Annotations}
An expert ASL interpreter involved in our project annotated a nearly 500-video subset from ASL STEM Wiki, which had previously been used for fingerspelling detection but did not contain glosses or other signing elements. 
The interpreter used the ASL video as the primary source and read the original English text as a secondary source. 
They annotated each sign, including glosses, fingerspelling, and classifiers. 
The first pass took an average of around 5 minutes to annotate each $\sim$15 second video from scratch. 
An extensive second pass was taken to create consistency across the annotation set, for example, ensuring the spelling of numbers and proper nouns was consistent, adding name sign labels, and correcting typos. 
For pseudo-annotation experiments, we used the Porter stemmer in NLTK~\cite{nltk} to normalize each word to prevent minor errors with suffixes (\eg DAYS instead of DAY).

In total there were 8,655 sign annotations, averaging about 17 per video. 
Of these, 6,201 are glosses, 1,969 are fingerspelled words, and 485 are classifiers. 
Of the glosses, 4,520 were in the ASL Citizen dictionary (411 unique) and the remaining 1,681 were not in this dictionary (355 unique). 
See the Supplemental Material for our interpreter's subjective assessments of this data.

\subsection{Subtask Datasets}

\noindent\textbf{Fingerspelling}: 
FSBoard~\cite{FSBOARD} is the largest public fingerspelling recognition dataset, which includes over 250 hours of 147 participants spelling short phrases, street addresses, numbers, and website addresses. 
Video were recorded from a smartphone ``in the wild'' and annotated at the video-level with the intended phrase. 
The vocabulary includes 26 English letters, numbers 0 to 9, and special characters. 

\noindent\textbf{Isolated Sign Recognition}: 
ASL Citizen~\cite{desai2023asl} is a large isolated sign language dataset consisting of 2,731 unique signs chosen from the ASL-LEX\cite{ASLLEX} vocabulary set. 
Each sign was recorded by up to 52 signers on their own using a device like a smartphone and totals over 83k videos. 
Contributors, primarily native Deaf signers, recorded themselves signing individual signs. Each video is labeled with its sign. 


\section{Experiments \& Discussion}
\label{sec:experiments}

First,  we evaluate and characterize performance on each intermediate task, demonstrating state-of-the-art performance on fingerspelling and ISR datasets. 
Second, we show that glosses can be an effective mechanism for translating ASL to English, as demonstrated using our gloss annotations. 
Lastly, we share statistics on the combined intermediate tasks as part of the full pseudo-annotations pipeline.

\subsection{Candidate Gloss Translations}
The following summarizes the LLM-based English-to-ASL Gloss translations using Claude Sonnet 4.5.
On ASL STEM Wiki (64,266 phrases), the K-Shot LLM with $k=10$ generated 7.3M glosses averaging 11.4 per sentence, with 45\% in the ASL Citizen vocabulary.
On FLEURS-ASL (1,749 phrases), 204k glosses were generated averaging 11.7 per sentence, with 53\% in-vocabulary.

The LLM under-predicts both fingerspelling and classifiers relative to our manual annotations: 13\% of predicted signs are fingerspelled versus 21.9\% in annotations, and classifiers account for only 0.1\% versus 5.4\%.
The LLM tends not to label signs as fingerspelled when there is ambiguity (\eg \texttt{SPECIES} was fingerspelled 91\% of the time in annotations but only 1.3\% in predictions).

Comparing LLM candidates to our manual annotations on ASL STEM Wiki, 82\% of annotated fingerspelled words and 78.5\% of in-vocabulary glosses appear in at least one candidate.
LLM glossings average 11.25 signs per phrase versus 13.1 in annotations, with common omissions being function words (\eg \texttt{FOR}, \texttt{THAT}).

\begin{table}[t]
	\centering
	\begin{tabular}{@{}lc}
		\toprule
		Model & Character Error Rate$\downarrow$ \\
		\midrule
		Kaggle Competition Winner \cite{chow2023google} & 16.4 \\
		ByT5~\cite{FSBOARD} & 11.1 \\
		ByT5 with YouTube-ASL~\cite{tanzer2024fingerspellingsignlanguagetranslation} & 8.9* \\
		Ours (no LLM) & 7.34 \\
		Ours (with LLM) & \textbf{6.75}  \\
		\bottomrule
	\end{tabular}
	\caption{Character Error Rate (CER) on the FSBoard dataset. Lower is better. *=Uses 2,800 hours of pretraining}
	\label{table:fingerspelling}
	\centering
    \setlength{\tabcolsep}{3pt}
	\begin{tabular}{l|l}
		\hline
		\textbf{Source} & \textbf{Output}   \\
		\hline
		Truth &  \texttt {29 OLD MOUNT PLEASANT} \\
		Ours & \texttt{29 OLD MOUNT \textcolor{red}{PLESANT}}\\
		Ours+LLM & \texttt{29 OLD MOUNT PLEASANT}\\
		\hline								
		Truth &  \texttt{HTTP://WWW.VISITDALARNA.SE} \\
		Ours  & \texttt{HTTP\textcolor{red}{NN}//W\textcolor{red}{W.}.VISI\textcolor{red}{S}TDALARNA\textcolor{red}{SE}E} \\
		Ours+LLM& \texttt{HTTP://WWW.VISITDALARNA.SE} \\
		\hline
		Truth &  \texttt{6408 NORTHEAST 364TH STREET} \\
		Ours  & \texttt{6408 \textcolor{red}{L} NORTHEAST \textcolor{red}{364 TH} STREET} \\
		Ours+LLM & \texttt{6408 \textcolor{red}{NE} 364TH STREET}\\
		\hline		
		
	\end{tabular}
	\caption{Fingerspelling examples from FSBoard with predictions generated by our base model and those generated with our base model with an error-correcting LLM on top using Claude 4.5.}
	\label{tab:fingerspelling_examples}
\end{table}

\subsection{Fingerspelling}


Table~\ref{table:fingerspelling} compares our fingerspelling recognizer with existing state-of-the-art methods on the FSBoard test set. 
We report the Character Error Rate (CER), which is calculated as the sum of substitutions, deletions, and insertions divided by the total number of characters in the ground truth. Lower CER indicates better performance. 
Our model with a simple greedy decoder achieves a CER of 7.3 and ours with LLM error correction achieves 6.75. 
Both variants perform better than competing models, despite the fact that the state of the art was trained with on over 1000 hours of other signing data~\cite{tanzer2024fingerspellingsignlanguagetranslation}.
The greedy decoder can be run in approximately real-time. The LLM on top takes $\sim$1 second after decoding. 

Table \ref{tab:fingerspelling_examples} shows examples from the dataset along with predictions from our base model and the model with the LLM on top. 
Two trends we see are that the LLM fixes many typos and missed letters, for example, when people slide their hand to repeat characters (\eg \texttt{WWW} in a website URL may look like \texttt{W-slide}). 
Interestingly, even when we asked the LLM not to make significant changes to the input, it would at times abbreviate text that was spelled out (\eg \texttt{NORTHEAST} to \texttt{NE}). 



Next we compare performance on video with a mix of fingerspelling and signing.
Our pseudo-annotation datasets contain a mixture of fingerspelling, glosses, and classifiers. 
We evaluate our model's performance on detecting whether or not (a) any word from the original English sentence or (b) any word from the annotated glosses were fingerspelled. We compute precision, recall, and area under the curve (AUC) when given the English or glossed text and ASL video, using the alignment approach described in Section~\ref{sec:fingerspelling_model}. 
We do not tune the model on this data. 
For the annotated portion of ASL STEM Wiki using English as input, the AUC is 0.800 and at the operating point of 0.3, which has the maximal F1 score, the precision is 0.892 and recall is 0.790. 
We find a large discrepancy between short and long words, which are common with abbreviations and chemical names in this dataset.
If we only evaluate on words that are more than 3 letters long AUC is 0.885. 
In the oracle case where we use annotated glosses as input, AUC overall is 0.897 and the AUC with words longer than 3 letters is 0.954. 
This suggests that if we predict the correct gloss predictions with the LLM and use them as input, our model should correctly detect whether or not the indicated words were fingerspelled or not. 
On both ASL STEM Wiki and FLEURS-ASL, many examples of individual letters as a middle initial in a name, chemical abbreviations (\eg \texttt{$R+$}), or math terms. 
It is conceivable that the number of variations in fingerspelling for this domain is higher than in other conversational domains.

\noindent\textbf{Temporal Boundary Evaluation:}
To evaluate the temporal accuracy of our fingerspelling model, we manually annotated start and end timestamps for 2,023 fingerspelled words in our 500-video annotated subset of ASL STEM Wiki.
A prediction is considered correct if it temporally overlaps with the corresponding annotation.
Our best model achieves 83.2\% F1 across all words and 88.9\% F1 on words with three or more characters.
Importantly, the model has not seen this dataset at training time, demonstrating strong generalization.

\subsection{Isolated Sign Recognition (ISR)}

Following past work, we report Top-1, Top-5, and Top-10 class accuracy when classifying a video as belonging to one of the 2,731 ASL Citizen classes. 
When incorporating handedness into our model, we achieve a 74\% Top-1 accuracy, whereas without handedness, we obtain 69\% accuracy. 
See Table~\ref{tab:islr_results} for all results.
The Top-5 and Top-10 performance is encouraging for our use cases in continuous signing video, because it means that even if a given class does not have the highest probability, the model is generally able to give a higher probability to the correct class. 


\begin{table}[t!]
	\centering
	\begin{tabular}{l|ccc}
		\toprule
		Model & R@1$\uparrow$ & R@5$\uparrow$ & R@10$\uparrow$ \\
		\midrule
		Baseline~\cite{desai2023asl} & 0.63 & 0.86 & 0.91 \\
		SignClip~\cite{zheng2023signclip} & 0.60 & 0.84 & 0.89 \\
		SHuBERT~\cite{gueuwou2025shubert} & 0.65 & 0.87 & 0.91\\
		Ours & 0.69 & 0.89 & 0.92 \\
		Ours (handed) & \textbf{0.74} & \textbf{0.91} & \textbf{0.94} \\
		\bottomrule
	\end{tabular}
	\caption{Isolated SL Recognition on ASL Citizen. R@k means recall for the top 1, 5, or 10 probability classes for a given video.}
	\label{tab:islr_results}
\end{table}

{\setlength{\fboxsep}{0pt}
\definecolor{mygreen}{rgb}{0.4, 0.9, 0.4}
\begin{figure*}[h]
    \centering
    \includegraphics[width=0.99\textwidth]{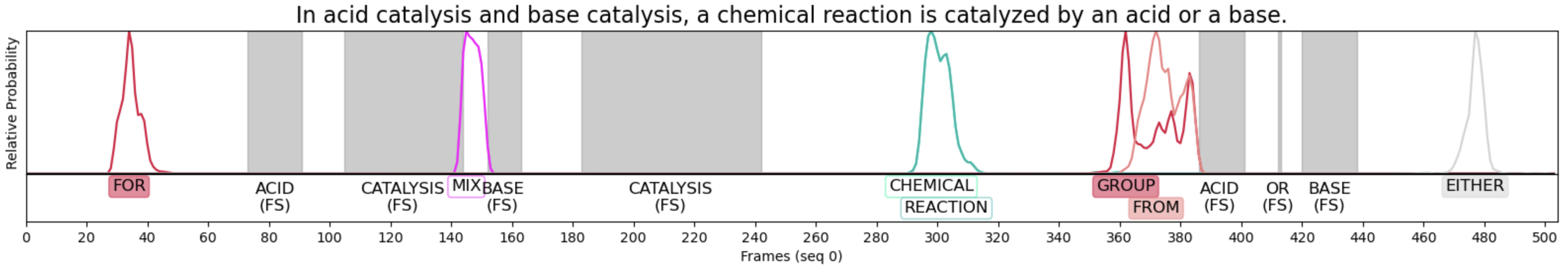}
    \includegraphics[width=0.99\textwidth]{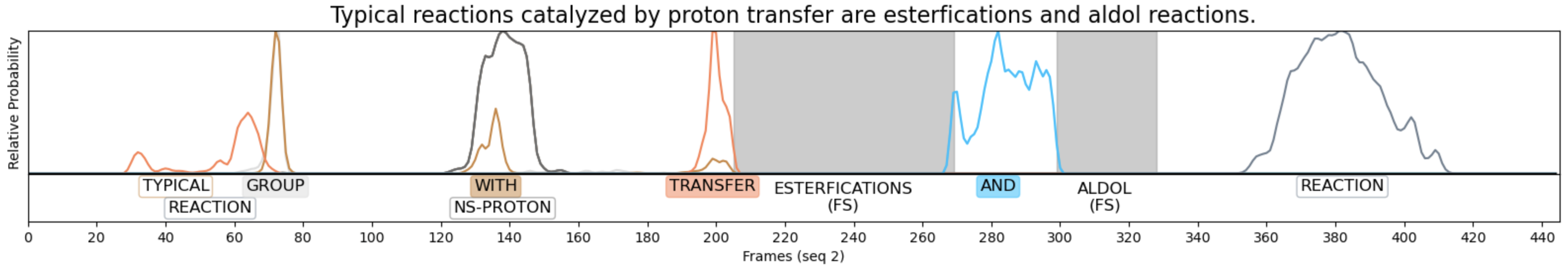}
    \caption{ Candidate forced alignment plots using our pseudo-annotation pipeline. Gray areas are \fcolorbox{gray}{gray}{\color{white}{fingerspelled}} regions as defined by detections of the first and last letter in a word. Gloss detections with a solid background are \setlength{\fboxrule}{0pt}\fcolorbox{mygreen}{mygreen}{in-vocabulary} and glosses that are outlined are  \setlength{\fboxrule}{1pt}\fcolorbox{blue}{white}{out-of-vocabulary} detections. Each colored line corresponds to the score of each gloss class. \texttt{(fs)} refers to a fingerspelled word.
    }
    \label{fig:timelines}
\end{figure*}

\begin{table*}[t]
	\centering
    \setlength{\tabcolsep}{3pt}
	\begin{tabular}{l|l}
		\toprule
		\textbf{Source} & \textbf{Output}   \\
		\midrule
English & At 23, the Kiev grid controller allowed the reactor shutdown to resume. \\
Annotation & FS-23 FS-KIEV SPREADSHEET CONTROL+PERSON ALLOW REACT SHUT-DOWN \\
Pipeline & AGE 23 FS-KIEV CONTROLLER GRID ALLOW REACTOR SHUTDOWN START+AGAIN \\
\midrule
English & The commercial banana is an example of a sterile, seedless triploid hybrid. \\
Annotation & BANANA THOSE BUSINESS MAKE EXAMPLE FS-STERILE FS-TRIPLOID FS-HYBRID  \\
Pipeline & BANANA COMMERCIAL SHOW EXAMPLE FS-STERILE SEEDLESS FS-TRIPLOID FS-HYBRID \\
\bottomrule    
	\end{tabular}
	\caption{Examples of English input, annotated input, and the best scoring translation.}
	\label{tab:pipeline_examples}
\end{table*}

\subsection{Pipeline \& Annotation Experiments}

A critical question that needs to be addressed is as follows: If we were able to accurately recognize signs using our annotation scheme, how accurately could we translate those to English? 
We use Claude Sonnet 4.5 using a gloss-to-English variation of the prompt in Section~\ref{sec:pipeline} to translate from our manual ASL STEM Wiki annotations to the reference English text that signers interpreted. 
The full prompt is in the supplemental material.
We align with recent work in the NLP community and use BLEURT~\cite{BLEURT}, ChrF~\cite{ChrF}, and Comet\footnote{Comet$_{22}$ version \texttt{Unbabel/wmt22-comet-da}}~\cite{Comet} to compare the original and predicted English phrases. 
See the supplemental material on why we intentionally do not use BLEU, despite it being common within the sign language recognition community.
The score for ChrF is 0.589, BLEURT is 0.612, and Comet is 0.715. All of these are out of 1.0, with higher being better. 
While these are very positive scores, note that these translations focus on STEM topics from Wikipedia, so further work is necessary to validate results on other domains.

While it is hard to quantitatively evaluate the final model scores, we ran an oracle experiment with both LLM-based candidate glosses and the annotated glosses and ranked the outputs. In 97\% of cases, the annotated glosses had the highest score across candidates. 
Errors were predominantly in cases where a sentence does not have any fingerspelling.

Figure~\ref{fig:timelines} visualizes example outputs from our system using candidate glossings. Gray intervals correspond to fingerspelling detections, and each colored line corresponds to a normalized score for a given gloss. Gloss names are listed underneath, where names with a background color are in-vocabulary, without a background color are out-of-vocabulary, and with ``(FS)'' notation are fingerspelled. 
Table~\ref{tab:pipeline_examples} shows example phrases, annotations and highest-scoring predictions. 
In the first example, glosses including \texttt{BANANA} and \texttt{EXAMPLE} are in-vocabulary and consistent between annotations and predictions. 
In the second example, the interpreter accidentally signed the word \texttt{SPREADSHEET}, which is out-of-vocabulary, instead of \texttt{GRID}, which is also out-of-vocabulary. 
This is an interesting case where the model correctly predicted that there were signs, but the annotations are inherently incorrect.

\section{Conclusion}
\label{sec:conclusion}

In this paper, we have made progress on our goal of scaling up annotation of sign language data through the development of pseudo-annotations.
This work should enable new efforts towards sign language understanding and recognition on the ASL STEM Wiki and FLEURS-ASL datasets. 
In addition, the ideas presented in this work could be used to generate annotations for additional datasets and be applied to other domains. 
Our approaches for fingerspelling recognition and isolated sign recognition can be trained with modest GPU resources and could also be used for further iteration on pseudo annotation pipelines. 
While automated sign language recognition continues to be a challenging problem, we hope that the manual annotations and pseudo-labels can be leveraged to make continued progress. 

\subsection{Acknowledgements}
Thank you to Gus Shitama, Julia Sohnen, and Lilian de Greef for numerous suggestions and recommendations on this work.  

{\small
\bibliographystyle{ieee_fullname}
\bibliography{egbib}
}

\newpage
\appendix
\onecolumn

\section{LLM Prompts}
\label{appendix:prompts}

In the paper we share results using three LLM prompts: (1) K-Shot English to ASL Gloss, (2) Fingerspelling: Error correction, and (3) Oracle back-translation: Manual Gloss Annotations to English. The full prompts are below.

\begin{llmprompt}{Direct Translation (K-Shot English to Gloss)}
You are an expert American Sign Language (ASL) translator. Your task is to translate English sentences into ASL glosses using standard ASL linguistic notation. You may be asked to generate more than one candidate translation.

OUTPUT FORMAT:\\
- Provide ONLY the ASL gloss translation\\
- No explanations, notes, definitions, or commentary\\
- No punctuation marks\\
- If asked to translate multiple translations, each translation should be enumerated with the dictionary structure below.\\
- Use ALL CAPS for signs\\

ASL GLOSS RULES: \\
1. NUMBERS: \\
- Write numbers 1-9 as digits: 1, 2, 3, etc.\\
- Numbers 10+ use conceptual signing: "100" → "1 HUNDRED", "25" → "2 5" or "TWENTY-FIVE"\\
- Years: Use full digits "1998" or conceptual "NINETEEN NINETY-EIGHT"\\
- Ages: "AGE" + number\\
- Time: Use appropriate time markers (MORNING, AFTERNOON, etc.)\\

2. GRAMMAR MARKERS:\\
- Use hyphens for compound signs: SELF-CONTROL\\
- Fingerspelling: JOHN for names not having signs. Don't use "FS-" or have dashes between letters.\\
- Use classifiers when appropriate (e.g., CL:1(POINT) or CL:4(list))\\
- Prefer classifiers instead of index notation\\
- Use a special character instead of text (e.g., use "+" instead of "PLUS")\\

3. ASL STRUCTURE: \\
- Every sentence you generate should consider using a different grammatical structure. \\
- Sometimes follow ASL word order (typically Time-Subject-Object-Verb)\\

4. COMMON CONVENTIONS:\\
- Past tense: FINISH or time markers, not English -ed\\
- Questions: Use question markers and facial expressions\\
- Negation: Use NOT before the sign \\

EXAMPLE:\\
English: "I am happy"\\
Output:\\
```json\\
{{\\
"1": "I AM HAPPY",\\
"2": "CL:1(point) AM HAPPY",\\
"3": I VERY HAPPY",\\
"4": "ME AM HAPPY",\\
"5": "I SO HAPPY",\\
"6": "I REALLY HAPPY",\\
"7": "I AM GLAD",\\
"8": "I FEEL GOOD",\\
"9": "ME FEEL HAPPY",\\
"10": "I AM JOY"\\
}}\\
\\
Please generate \{k\} different gloss translations \\
\\
NOW COMPLETE:\\
English: "\textit{\{phrase\}}"\\
Output: \\
```\\
\end{llmprompt}

\begin{llmprompt}{Fingerspelling: Error correcting prompt}
Your task is to take a line of english text and fix typos.
The text may contain a name, a physical address, a phone number, or a website address.
First think about if it is a website, phone number, name, or home address. Then think about how it might be wrong and how to improve. \\
\\
RULES: \\
 - The input may be correct as-is or it may need to be fixed. \\
 - Sometimes websites are only the terms in slashes like '/home/site/test/' or 'home-site-test'. Websites should have no spaces. \\
 - Phone numbers generally have dashes between sets of digits. \\
 - Do not remove full words or add new words. \\
 - Do not abbreviate names. Do not add commas. Do not remove any numbers.  \\
 - Respond with a string response only with nothing else. Do not wrap with json. Do not add notes.
 \\
 \\
NOW COMPLETE:\\
English: "\textit{\{phrase\}}"\\
Output: \\
\end{llmprompt}

\begin{llmprompt}{Oracle Experiment: Manual gloss annotations to English (back translation)}
You are an American Sign Language translator.
    Your task is to translate ASL glosses into complete English sentences.
    No notes, definitions, or anything other than glosses.
    Grammar should be in standard English word order.
    You may see other unicode symbols so translate those as you see fit.
    Words annotated with 'FS-' at the beginning are fingerspelled and should be replaced with the same word in English.
    Respond with a string response only with nothing else. Do not add notes.
\\
\\
NOW COMPLETE:\\
Glosses: "\textit{\{phrase\}}"\\
Output: \\
\end{llmprompt}

\section{Limitations of BLEU for Sign Language Translation}
\label{appendix:metrics}

Traditional machine translation metrics like BLEU~\cite{papineni_bleu_2002}, which rely on n-gram overlap between predicted and reference translations, prove inadequate for evaluating sign language translation systems. BLEU's focus on exact lexical matches fails to capture the semantic equivalence that is crucial in sign language translation, where multiple valid translations can express the same meaning using different vocabulary, word order, or levels of detail. This limitation is particularly pronounced in sign language translation due to the inherent differences between signed and spoken language modalities, including more flexible word order, frequent omission of function words, and the use of spatial and visual elements that may not have direct textual equivalents.

Our toy examples, shown in Table~\ref{tab:bleu}, demonstrate clear cases where BLEU scores are misleading or counterintuitive. All metrics range from 0 to 1 where higher is better. For instance, a translation that captures the core semantic meaning but uses synonyms or slightly different phrasing receives a low BLEU score despite being perfectly acceptable to human evaluators. Conversely, translations that maintain surface-level similarity through exact word matches but miss crucial semantic nuances can achieve artificially high BLEU scores while being rated poorly by human judges. These examples highlight BLEU's inability to account for semantic equivalence, paraphrasing, and the contextual appropriateness that human evaluators naturally consider when assessing translation quality.

\begin{table}[b]
    \centering
    \begin{tabular}{l|l|cccc}
         Type & Phrase  \\
         \toprule
         English Reference & I will go to the store tomorrow. & BLEU & ChrF & BLEURT & Comet$_{22}$ \\
         \hline
         ASL-like glosses & TOMORROW STORE TRAVEL I \\
         ASL-like back translation & Tomorrow I will travel to the store. & 27.05 & 59.72 & 0.767 & 0.84 \\
         SEE-like glosses & I TRAVEL STORE TOMORROW \\
         SEE-like back translation & I will travel to the store tomorrow. & 59.46 & 77.58 & 0.795 & 0.89 \\
         \toprule
         English Reference & The window was broken by the baseball. & BLEU & ChrF & BLEURT & Comet$_{22}$ \\
         \hline
         ASL-like glosses & BASEBALL HIT WINDOW, WINDOW BREAK \\
         ASL-like back translation & The baseball hit the window and the window broke. & 6.27 & 54.30 & 0.681 & 0.79 \\
         SEE-like glosses & WINDOW BROKEN FROM BASEBALL \\
         SEE-like back translation & The window broke from the baseball & 26.65 & 62.75 & 0.719 & 0.81 \\
         \bottomrule
    \end{tabular}
    \caption{Results across NLP metrics when back-translating an English reference with ASL-like glosses and Signed English (SEE)-like glosses. All back-translations accurately represent the English reference and thus are high quality, but scores for BLEU are very different, ChrF metrics are somewhat different, and both Comet and BLEURT are similar. }
    \label{tab:bleu}
\end{table}

In contrast, learned metrics such as COMET~\cite{Comet} and BLEURT~\cite{BLEURT}, which leverage pre-trained language models to capture semantic similarity rather than surface-level overlap, show significantly better alignment with human judgment in our evaluation. COMET's ability to recognize semantic equivalence between different lexical realizations makes it particularly well-suited for sign language translation, where the goal is faithful meaning transfer rather than literal word-for-word correspondence. The superior correlation between COMET scores and human ratings across our test cases suggests that the field should move toward adopting these more sophisticated evaluation metrics.

ChrF~\cite{ChrF} is also a popular machine translation metric and is a reasonable middle ground between BLEU and COMET. Unlike BLEU's strict n-gram matching, ChrF operates at the character level and uses F-score rather than precision-based scoring, making it more robust to morphological variations and word order differences that are common in sign language translation. While ChrF still relies on surface-level string matching rather than semantic understanding, its character-based approach allows it to capture partial matches and morphological similarities that BLEU would miss entirely. In our evaluation, ChrF scores show moderate correlation with human judgment—better than BLEU but not as strong as COMET—reflecting its position as an improved lexical metric that nonetheless lacks the semantic awareness necessary for optimal sign language translation evaluation.

These findings have important implications for sign language translation research, as the choice of evaluation metric directly influences model development, comparison of different approaches, and assessment of progress in the field. We recommend that future work in sign language translation prioritize semantic-aware metrics like COMET over traditional n-gram based metrics, and encourage the development of specialized evaluation frameworks that account for the unique linguistic properties of sign languages and their translation into spoken language text.

\section{Architecture Comparison: TCN vs.\ Conformer}
\label{appendix:architecture}

We selected Temporal Convolutional Networks (TCNs) for their fixed receptive field, hypothesizing this would improve cross-dataset generalization compared to attention-based architectures.
Post-submission experiments with a Conformer validate this choice.
As shown in Table~\ref{tab:model_comparison_supp}, while the Conformer achieves better in-domain results on FSBoard (which contains only fingerspelling), it overfits and performs significantly worse on out-of-domain data from ASL STEM Wiki and FLEURS-ASL, which contain a mixture of signing and fingerspelling.
For a system designed to pseudo-annotate diverse real-world signing data, the TCN's superior generalization is the preferred engineering choice.

\begin{table}[h]
\centering
\begin{tabular}{l|cc|cc}
\hline
\textbf{Dataset}  & \multicolumn{2}{c|}{\textbf{Conformer}} & \multicolumn{2}{c}{\textbf{TCN}} \\
                 & All & Long & All & Long \\
\hline
FSBoard (CER $\downarrow$) & \textbf{4.9\%} & -- & 7.3\% & -- \\
ASL STEM Wiki (AUC $\uparrow$) & 0.70 & 0.79 & \textbf{0.80} & \textbf{0.89} \\
FLEURS-ASL (AUC $\uparrow$) & 0.69 & 0.73 & \textbf{0.72} & \textbf{0.76} \\
\hline
\end{tabular}
\caption{Fingerspelling results comparing Conformer and TCN architectures. FSBoard is in-domain (only fingerspelling); ASL STEM Wiki and FLEURS-ASL are out-of-domain (signing \& fingerspelling). ``Long'' refers to words with $\geq$ 3 characters.}
\label{tab:model_comparison_supp}
\end{table}

\section{LLM Ablation Studies}
\label{appendix:llm_ablations}

\subsection{Model Selection}
We performed extensive experiments with different LLMs for gloss-to-English back-translation on ASL STEM Wiki.
Table~\ref{tab:llm_selection} shows BLEURT scores when translating from annotated ASL glosses to English.
Larger models tend to perform better, but there is low variance between frontier models.

\begin{table}[h]
\centering
\begin{tabular}{l|c}
\hline
\textbf{Model} & \textbf{BLEURT} $\uparrow$ \\
\hline
Gemma-3:12b & 0.52 \\
Gemma-3:27b & 0.55 \\
Claude Sonnet 4.0 & 0.59 \\
Gemini-2.5-flash & 0.60 \\
\hline
\end{tabular}
\caption{BLEURT scores for different LLMs on gloss-to-English back-translation using ASL STEM Wiki manual annotations.}
\label{tab:llm_selection}
\end{table}

\subsection{K-Shot Prompting Ablation}
There are many valid English-to-ASL translations, including some with very different grammar.
We validated the K-shot approach by measuring ChrF when comparing manual annotations to the best-case LLM translation for $K=1$ to $K=10$.
As shown in Table~\ref{tab:kshot}, gloss-level ChrF improves from 47.7 ($K=1$) to 50.8 ($K=10$), with 67\% of the improvement achieved by $K=4$.
This improvement is consistent for shorter and longer phrases.

\begin{table}[h]
\centering
\begin{tabular}{l|c}
\hline
$K$ & \textbf{Gloss-level ChrF} $\uparrow$ \\
\hline
1 & 47.7 \\
4 & 49.8 \\
10 & 50.8 \\
\hline
\end{tabular}
\caption{Effect of K-shot prompting on gloss-level ChrF, comparing manual annotations to the best-case LLM translation.}
\label{tab:kshot}
\end{table}

\section{Summary of ASL STEM Wiki Interpretations}
\label{appendix:interpretations}
There were 16 unique signers across the videos annotated.
The following summarizes the interpreters subjective assessments.
The signers in ASL STEM Wiki can be generally separated into what appears to be native or native-like ASL users and second-language (L2) learners.
Strong ASL users (Subjects 3, 4, 9, 11, 12, and 14) demonstrate proper ASL grammar structure, effective use of classifiers and depiction, clear fingerspelling, and appropriate use of non-manual markers.
These signers also tend to avoid unnecessary fingerspelling when established signs exist and show fluent, natural signing patterns without hesitation.
These signers, particularly Subjects 3 and 4, represent the gold standard for ASL output, with Subject 4 showing strategies typical of Certified Deaf Interpreters who aim for maximum comprehension across diverse audiences.

In contrast, several signers (Subjects 1, 6, 7, 10, 13, 15, and 16) exhibit what the Deaf community describes as a ``hearing accent,'' which are characteristics typical of second language (L2) ASL learners~\cite{pichler2011sources}.
Such signers show varying degrees of English influence, from Subject 1's heavy reliance on Signing Exact English (SEE) and fingerspelling, to Subject 10's less precise fingerspelling style and sign initialization.
These L2 signers typically follow English word order, have limited classifier use, and may hesitate or re-start mid-sentence while processing their interpretations.
Some L2 signers also adopt textbook-style signing that lacks the natural flow of native users and employ unnecessary or less precise fingerspelling.

\end{document}